\documentclass[sigconf]{acmart}
\usepackage{times}
\usepackage{soul}
\usepackage{url}
\usepackage{subfigure}
\usepackage{amsmath, bm}
\usepackage{multirow}
\AtBeginDocument{%
  \providecommand\BibTeX{{%
    \normalfont B\kern-0.5em{\scshape i\kern-0.25em b}\kern-0.8em\TeX}}}

\setcopyright{acmcopyright}
\copyrightyear{2018}
\acmYear{2018}
\acmDOI{10.1145/1122445.1122456}

\acmConference[Woodstock '18]{Woodstock '18: ACM Symposium on Neural
  Gaze Detection}{June 03--05, 2018}{Woodstock, NY}
\acmBooktitle{Woodstock '18: ACM Symposium on Neural Gaze Detection,
  June 03--05, 2018, Woodstock, NY}
\acmPrice{15.00}
\acmISBN{978-1-4503-XXXX-X/18/06}

\begin{document}

\title{Fast Video Salient Object Detection via Spatiotemporal Knowledge Distillation}

\author{Yi Tang}
\authornotemark[1]
\email{ytang@ie.cuhk.edu.hk}
\affiliation{%
  \institution{Department of Information Engineering}
  \city{Hong Kong}
}

\author{Yuanman Li}
\email{yuanmanli@szu.edu.cn}
\affiliation{%
  \institution{Institution of Electronic and Information Engineering}
  \city{Shenzhen}
}
\author{Wenbin Zou}
\email{zouwb@szu.edu.cn}
\affiliation{%
  \institution{Institution of Electronic and Information Engineering}
  \city{Shenzhen}
}
%
%
%
%
%
%
%


\begin{abstract}
  Since the wide employment of deep learning frameworks in video salient object detection, the accuracy of the recent approaches has made stunning progress. These approaches mainly adopt the sequential modules, based on optical flow or recurrent neural network (RNN), to learn robust spatiotemporal features. These modules are effective but significantly increase the computational burden of the corresponding deep models. In this paper, to simplify the network and maintain the accuracy, we present a lightweight network tailored for video salient object detection through the spatiotemporal knowledge distillation. Specifically, in the spatial aspect, we combine a saliency guidance feature embedding structure and spatial knowledge distillation to refine the spatial features. In the temporal aspect, we propose a temporal knowledge distillation strategy, which allows the network to learn the robust temporal features through the infer-frame feature encoding and distilling information from adjacent frames. The experiments on widely used video datasets (e.g., DAVIS, DAVSOD, SegTrack-V2) prove that our approach achieves competitive performance. Furthermore, without the employment of the complex sequential modules, the proposed network can obtain high efficiency with 0.01s per frame.
\end{abstract}




\maketitle

\section{Introduction}
The purpose of salient object detection (SOD) is to focus on the most attractive objects or regions in an image or video
and then highlight them from the complex background. As the trait of this community, it is usually treated as the preprocessing to support other visual tasks, such as visual tracking, image retrieval and photo cropping. As for saliency detection, we can roughly divide it into two categories. The first one is saliency prediction, which is to estimate the coarse saliency regions by catching the viewers' eye movement. The second one is salient object detection, which is similar to segmentation. It needs to classify each pixel and highlight completely the fine-grained the salient objects. Additionally, according to the input of frameworks, SOD also can be categorized into image salient object detection (ISOD) and video salient object detection (VSOD). In this paper, our research emphasizes on the more complex topic, namely, video salient object detection.

\begin{figure}[t]
\centering
\subfigure[Consecutive frames.]{
\includegraphics[width=0.38\textwidth]{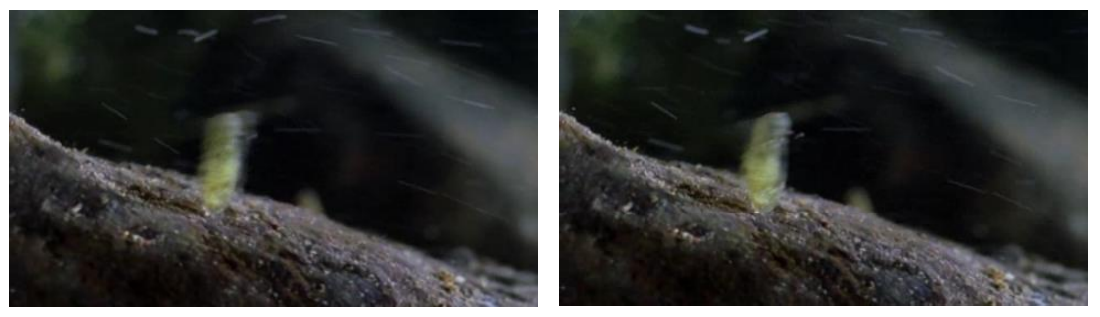}
}
\subfigure[Saliency maps from static saliency detector \protect\cite{deng2018r3net}.]{
\includegraphics[width=0.38\textwidth]{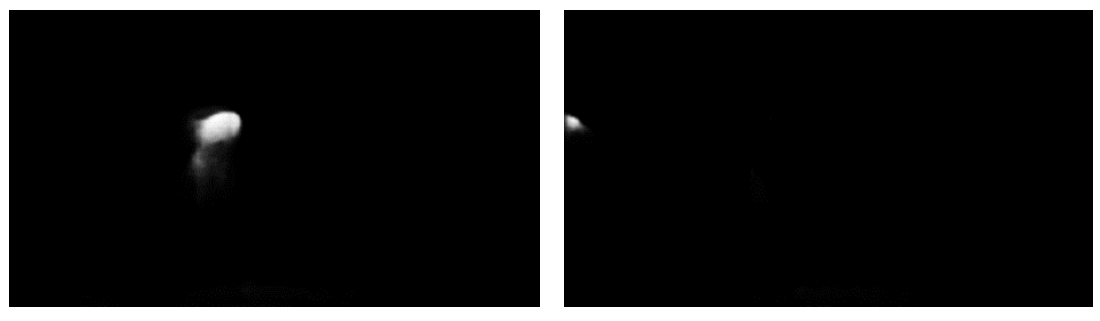}
}
\subfigure[Saliency maps from dynamic saliency detector \protect\cite{Fan_2019_CVPR}.]{
\includegraphics[width=0.38\textwidth]{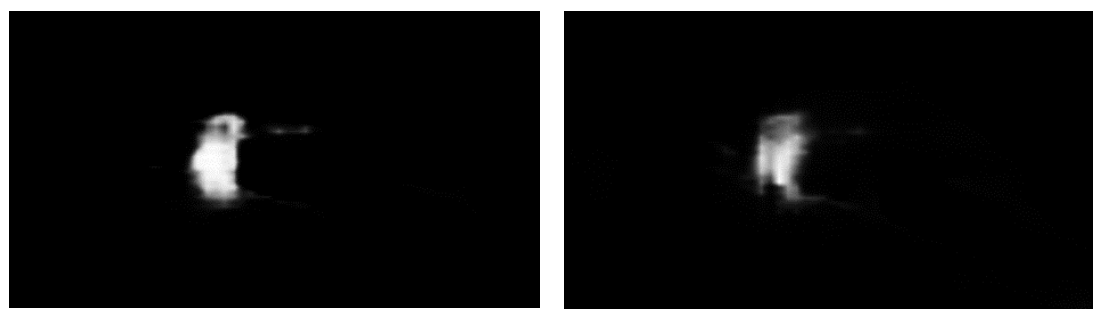}
}
\subfigure[Results from the proposed method.]{
\includegraphics[width=0.38\textwidth]{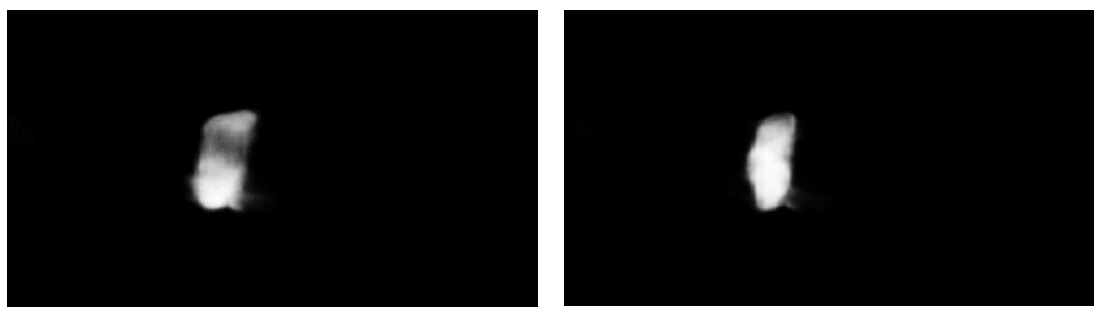}
}
\subfigure[Ground truths.]{
\includegraphics[width=0.38\textwidth]{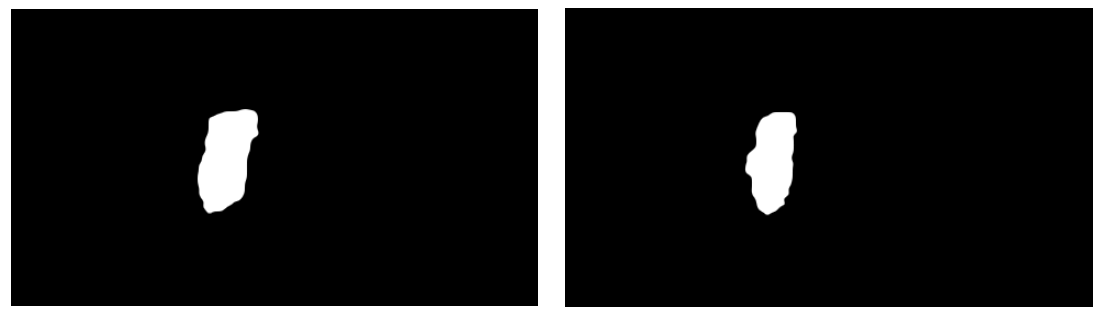}
}
\caption{Saliency results from different approaches.}
\label{pull_fig}
\vspace{-10px}
\end{figure}

Compared with the SOD in still images, moving objects are more likely to become the salient objects in videos. Therefore, inter-frame sequential information, that helps the algorithms capture the moving objects, plays a significant role to detect the salient objects in video scenes. As shown in Fig.\ref{pull_fig}, the image saliency detection method \cite{deng2018r3net} cannot capture the sequential information, thus missing the salient object in video sequences. Among the video saliency approaches, in order to obtain sequential information, researches try to build different dynamic models. Before the wide employment of deep models, the heuristic models \cite{wang2015consistent,chen2017video} are main solutions for this field. These kinds of methods build the spatiotemporal models by the hand-crafted features and optical flow-based motion cues, which cannot guarantee the performance and runtime of the algorithms. With the introduction of deep learning, the performance of VSOD has been significantly improved. However, as for some complex video scenes (e.g., low resolution, motion blur, scale change and so on), the early deep models \cite{8047320,8419765} are limited to extract robust sequential features by the straightforward fully convolutional network (FCN). Therefore, it is difficult for these dynamic models to further improve the performance in VSOD. In order to solve these difficulties, the current approaches \cite{li2018flow,8419765,song2018pyramid,Fan_2019_CVPR} try to append some complex sequential extractors to refine the spatiotemporal features. These specific sequential extractors mainly contain optical flow processing modules and RNN based modules. All of them are able to remarkably boost the performance, but they also make the deep models bigger and raise the computational cost at the same time. Especially for optical flow-based methods \cite{8419765,li2018flow}, the optical flow needs to be obtained by an extra deep model (e.g., FlowNet \cite{8099662}), which directly affects the efficiency of the entire saliency algorithm. Therefore, it is still a challenge to allow the video saliency model to learn robust sequential information in the lightweight condition.

Knowledge distillation is originally a model compression approach, whose structure includes two deep models, namely, a large teacher model and a small student model. The student model is used to mimic the teacher model so that it can obtain the significant performance of the teacher model. Along with this kind of pattern, many approaches \cite{liu2019structured,li2019spatiotemporal,zhang2019training} are able to obtain reliable and efficient deep models to conduct their tasks. However, traditional distillation needs to train two deep models, which spend more training time, thus directly reduce the efficiency of the entire approach. With the introduction of self distillation \cite{Zhang_2019_ICCV}, this issue can be gradually alleviated. The consideration of self distillation is to exploit the own generated results as the supervision to complete the knowledge transfer.


Inspired by the self distillation strategy, we propose a spatiotemporal knowledge distillation strategy to conduct video salient object detection. The main idea is to distill the information from adjacent frames, and then transfer the infer-frame information to the current frame. Specifically, the proposed spatiotemporal distillation mainly includes two components, namely the spatial distillation and temporal distillation. For the spatial distillation, we introduce a saliency guidance feature embedding module, in which the rear generated saliency maps are utilized as a supervision to propagate the distilling knowledge to the front hidden layers. For the temporal distillation, in order to allow the lightweight network to learn robust infer-frame information, we introduce an attention-based infer-frame feature encoding module, which integrates with the distilling information from adjacent frames to propagate the sequential information.

In summary, the main contributions of this paper can be concluded as below:

\begin{itemize}
  \item We propose a spatiotemporal knowledge distillation approach for the salient object detection in video scenes, which can effectively make the lightweight network obtain robust spatiotemporal features.
  \item We introduce an attention-based infer-frame feature encoding module, which combines with the temporal distillation to adequately propagate the sequential information from adjacent frames to the current frame.
  \item The experiments on the widely used datasets, such as DAVIS, DAVSOD, SegTrack-V2, demonstrate that the proposed approach achieves competitive performance against the state-of-the-arts. Furthermore, the runtime of the lightweight model is very fast with 0.01s per frame.
\end{itemize}

\section{Related Work}
\subsection{Video salient object detection}

With the introduction of deep learning methods, we can divide previous video saliency approaches into two phases, which are non-deep learning approaches and deep learning-based approaches. The formers are based on hand-crafted features (e.g., color contrast, texture, optical flow) and heuristic models (e.g., geodesic distance, center-surround contrast). As the exploitation of optical flow and complex optimization models, these approaches are usually time-consuming. Moreover, they also cannot handle some complex video scenes such as motion blur, low contrast, occlusion, because of the limitation of hand-crafted features. With the success of deep learning in many visual tasks, the community of VSOD also starts utilizing deep learning techniques widely \cite{8047320,Chen2018TIP,8419765,song2018pyramid,Fan_2019_CVPR}. In the beginning, as the shortage of training data, weakly supervised method \cite{8419765} semi-supervised method \cite{Yan_2019_ICCV} and data synthetic \cite{8047320} are employed to produce pixel-wise labels. In the aspect of deep saliency models, dilated convolutions and LSTM-based structures are introduced to retain sufficient feature scale and extract motion information, respectively. For example, PDB \cite{song2018pyramid} extracts spatial features and temporal features by pyramid dilated convolutions and deeper bi-directional ConvLSTM structure, respectively. Based on PDB, SSAV \cite{Fan_2019_CVPR} further introduces eye fixation records for network training and proposes a saliency-shift-aware ConvLSTM module to capture video saliency dynamics. Additionally, DLVS \cite{8047320} proposes a saliency detection network by stepwise extracting static and dynamic information. Li et al. \cite{li2018flow} subtly combines optical flow and short connection structure to propose a flow guided recurrent neural encoder framework. Along with the deep features, SCOM \cite{Chen2018TIP} builds a spatiotemporal constrained optimization model for VSOD. Recently, MGA \cite{Li_2019_ICCV} proposes an end-to-end network, which effectively integrates the spatial features from a single frame and the temporal features from the optical flow.

\begin{figure}[t]
\centering
\subfigure[The prediction of the teacher model and ground truth are regarded as the supervision to train the student model.]{
\includegraphics[width=0.45\textwidth]{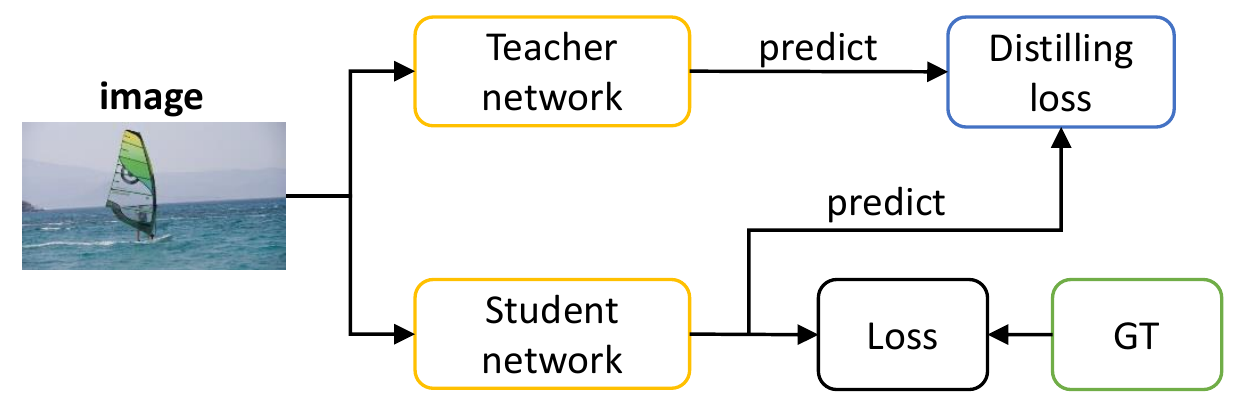}
}
\subfigure[The prediction from the deep layers and ground truth are regarded as the supervision to train the shallow layers. ]{
\includegraphics[width=0.45\textwidth]{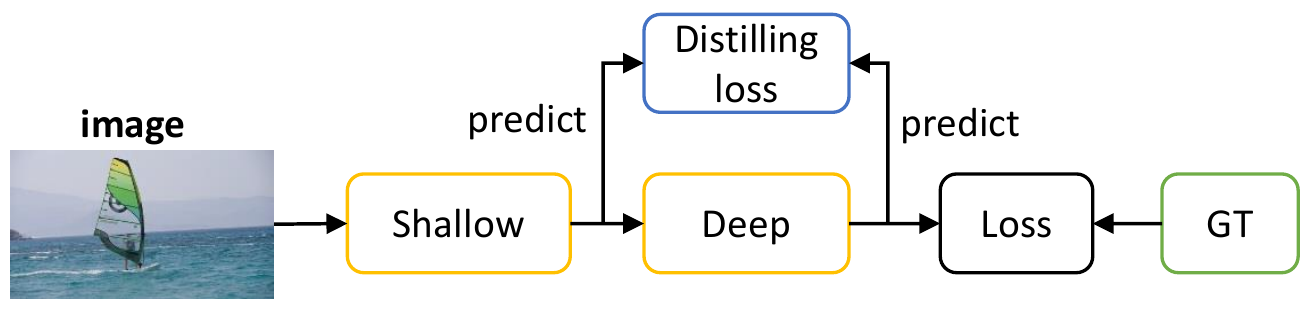}
}
\caption{A brief description of the traditional knowledge distillation and self-distillation. (a) The process of the traditional knowledge distillation. (b) The process of the self-distillation.}
\label{distill}
\end{figure}

\subsection{Knowledge distillation}

The idea of knowledge distillation is firstly proposed in \cite{bucilua2006model}. Then, Hinton et al. \cite{hinton2015distilling} further explain the concept of distillation. At first, knowledge distillation is referred to as the process that a teacher teaches a student the knowledge. In this processing, we have to train two deep models, namely a teacher model and a student model. The teacher model is a large model. The student model is a small model, which is trained by the strong labels from ground truths and the prediction from the teacher model. This process can be described in Fig.\ref{distill} (a). Relying on this strategy, Liu et al. propose a structured knowledge distillation \cite{liu2019structured}, which combines distillation and generative adversarial network (GAN) to conduct semantic segmentation. Additionally, knowledge distillation can also be applied for other domains, such as image classification \cite{li2017learning}, pedestrian re-identification \cite{chen2018darkrank} and saliency prediction \cite{zhang2019training}. Although the distillation mechanism can help the target model boost their performance, the knowledge transfer is of low efficiency with the two-stage training tasks. In order to solve this issue, researchers propose a self distillation strategy, which can be described as ``be your own teacher" \cite{Zhang_2019_ICCV}. Specifically, they treat the probability of the deep layers as the supervision to optimize the shallow hidden layers, which is shown in Fig.\ref{distill} (b). In \cite{Hou_2019_ICCV}, Hou et al. combine an attention mechanism and the self distillation for lane detection. As only one model needs to be trained, self distillation can dramatically improve training efficiency.

\subsection{Visual attention}

The attention mechanism is to compute the weights from different and complex information. It is usually used for selecting and fusing features in many applications, such as image caption \cite{chen2017sca}, image segmentation \cite{fu2019dual,yuan2018ocnet}, etc. In particular, the work \cite{yuan2018ocnet} proposes an object context pooling scheme by exploiting self-attention module, whose principle is to compute similarities of all pixels and them selectively integrates them. DANet \cite{fu2019dual} further proposes two kinds of attention modules: position attention module (PAM) and channel attention module (CAM). The two modules can explore the contextual dependencies in FCN framework. Different from their PAM based on self-attention, we build a mutual attention module to obtain the contextual information in spatial and temporal cues. In \cite{yang2019anchor}, Yang et al based on the self-attention module to propose a simple yet effective attention module to encode the infer-frame features for unsupervised video object segmentation.

\section{The proposed method}

\begin{figure*}[t]
\centering
\subfigure[The multi-level feature maps are extracted from the backbone network and ASPP module.]{
\includegraphics[width=0.9\textwidth]{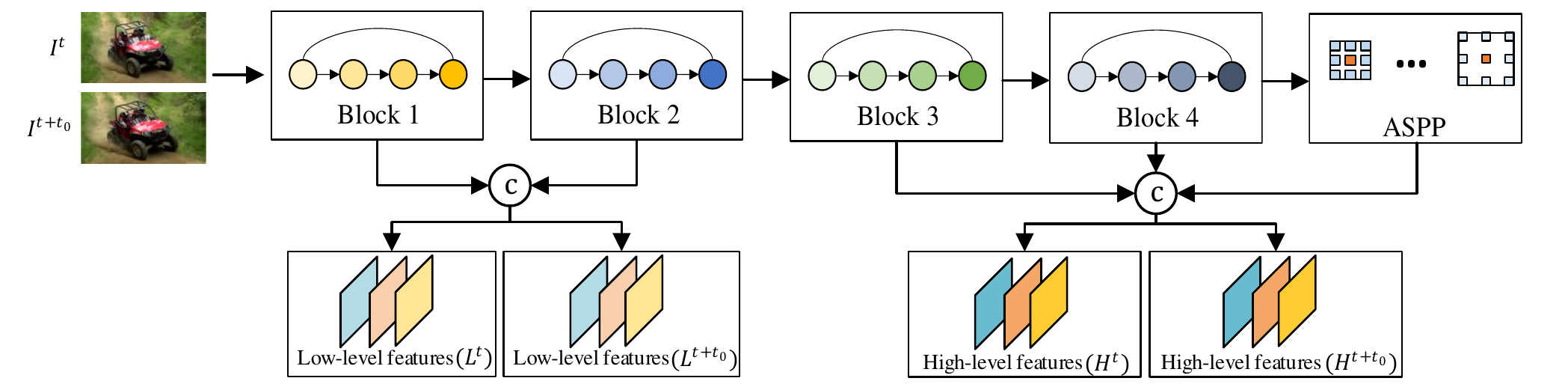}
}
\subfigure[The feature embedding units and spatial distillation are employed to refine the features in the spatial component. ]{
\includegraphics[width=0.9\textwidth]{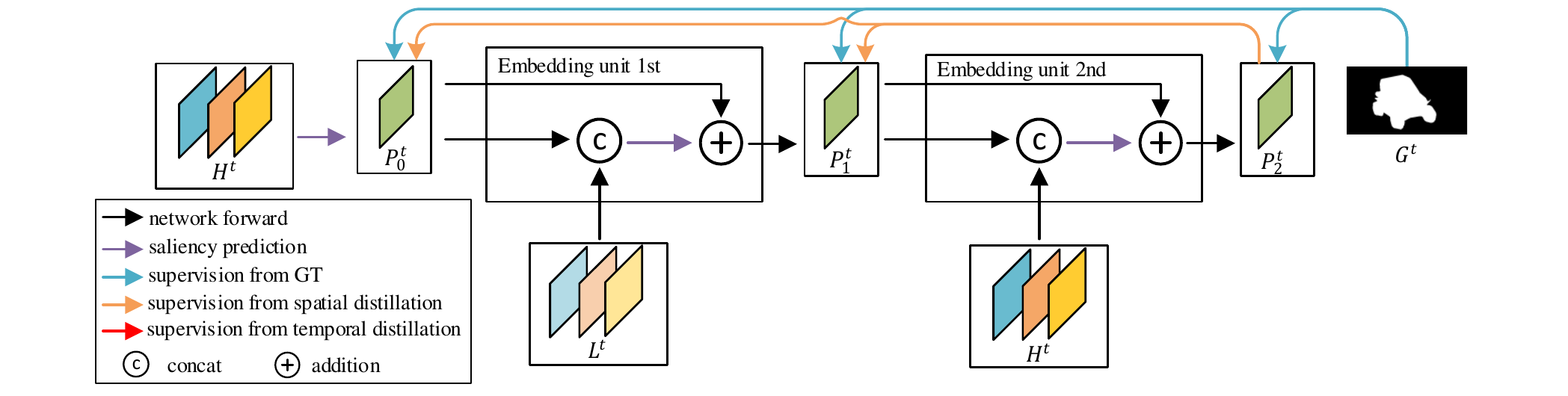}
}
\subfigure[An extra feature embedding unit is introduced to encode the infer-frame features. Along with the temporal distillation, the sequential information is effectively transferred from the adjacent frames to the current frame]{
\includegraphics[width=0.9\textwidth]{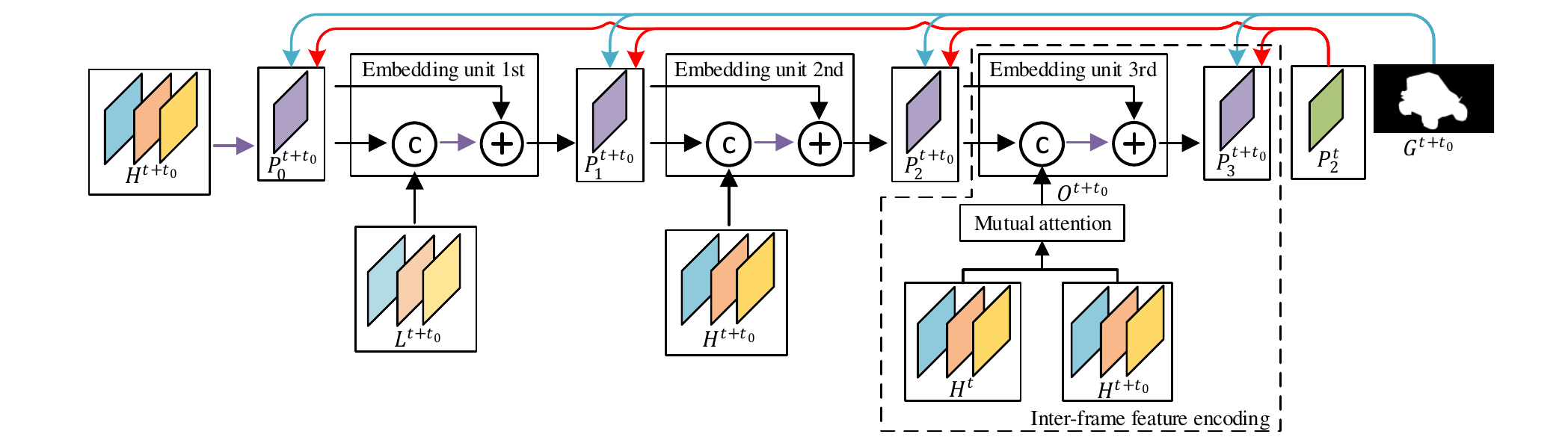}
}
\caption{The overall framework of the proposed spatiotemporal knowledge distillation. (a) The extraction of multi-level features. (b) Saliency guidance feature embedding and spatial distillation strategy. (c) Infer-frame feature encoding module and temporal distillation strategy.}
\label{framework}
\vspace{-10px}
\end{figure*}

\subsection{Overview}

As shown in Fig.\ref{framework}, the proposed lightweight network with the spatiotemporal knowledge distillation consists of three components. The first one is the feature extractor, which includes a backbone network and an atrous spatial pyramid pooling module (ASPP) \cite{chen2017rethinking}. Given the adjacent frames, the low-level and high-level features are extracted. In the second component, we introduce a saliency guidance feature embedding structure \cite{deng2018r3net} to encode the multi-level features. Then, the spatial knowledge distillation is employed to refine the spatial information. In the third component, an infer-frame feature encoding module and temporal knowledge distillation are simultaneously exploited to transfer the sequential information. 

\subsection{Multi-level feature extraction}

In our network, we utilize the ResNet-50 as the backbone and the ASPP module to refine the spatial features. The backbone is composed of four residual convolutional blocks. The ASPP module as the fifth block is a group of hierarchical dilated convolutional layers, which can further extract the robust spatial features by different receptive fields of convolution. Given the adjacent frames $I^{t}$, $I^{t+t_0}$ ($t_0 \in N^{*}, \|t_0\| \leq 3$) as the network input, the multi-level feature maps from the five blocks are extracted. Following the feature embedding method \cite{deng2018r3net}, the side outputs from the first two blocks are concatenated to generate the low-level features $L^{t}$, $L^{t+t_0}$. The outputs from the last three blocks are exploited to generate the high-level features $H^{t}$, $H^{t+t_0}$. The features extraction can be formulated as follows:

\begin{equation}
\label{extraction}
\begin{aligned}
   S^{t}_i  &= F(I^{t};\bm{W_s}), i=1, 2, ..., 5 \\
   L^{t}  &= \Gamma(Cat(S^{t}_1, S^{t}_2);\bm{W_l}) \\
   H^{t}  &= \Gamma(Cat(S^{t}_3, S^{t}_4, S^{t}_5);\bm{W_h}),
\end{aligned}
\end{equation} where $S^{t}_i$ is the side output, $i$ represents the convolutional block. $\Gamma(\cdot; \cdot)$ denotes the a group of convolutional operations. $Cat(\cdot)$ demonstrates the concatenation operation. $\bm{W_s}$, $\bm{W_l}$ and $\bm{W_h}$ are the corresponding model parameters.

\subsection{Spatial knowledge distillation}

After the extraction of the multi-level feature maps, we introduce a saliency guidance feature embedding structure from \cite{deng2018r3net}. Its purpose is to exploit the previous saliency map to guide the network learning, thus revising the errors in previous saliency maps. The specific pipeline is illustrated in Fig.\ref{framework} (b). The high-level feature maps $H^{t}$ are firstly used to generate the first phase saliency prediction $P^{t}_0$ with a convolutional layer. Secondly, the low-level feature maps $L^{t}$ are concatenated with the first phase saliency prediction $P^{t}_0$. Then, a group of convolutional operations $\Phi_j$ and an element-wise addition are exploited to generate the second phase saliency prediction $P^{t}_1$ in the feature embedding unit. Similarly, the third phase saliency prediction $P^{t}_2$ is generated by the $P^{t}_1$ and $H^{t}$. The forward propagation of the feature embedding units can be described as below:

\begin{equation}
\label{recurrent}
\begin{aligned}
   R^{t}_j  &= \Phi_j(Cat(P^t_{j-1}, M^t)) \\
   P^{t}_{j}  &= P^t_{j-1} \oplus R^{t}_j, j=1, 2,
\end{aligned}
\end{equation} where $\Phi_j$ is the convolutional operations, which include three convolution layers with batch normalization and PReLU activation. $M^t$ is the input of $\Phi_j$. $R^{t}_j$ represents the residual, which is added with the previous phase saliency prediction $P^t_{j-1}$ to generate the current prediction $P^{t}_{j}$. Different from \cite{deng2018r3net}, multiple embedding units are adopted to alternatively integrate the low-level and high-level features. In this paper, we hope to present a lightweight network, so two embedding units are sufficient to encode the multi-level features. Further, combined with the knowledge distillation, the adopted structure can obtain competitive performance as well.


In the backward propagation, the spatial knowledge distillation is also depicted in Fig.\ref{framework} (b). We treat the deepest section as the teacher model to transfer the knowledge to the swallow section, which is regarded as the student model. Here, the final phase saliency prediction $P^t_{2}$ as the supervision is employed to optimize the previous two saliency predictions $P^t_{0}$, $P^t_{1}$. Therefore, the loss function of the spatial distillation $L_s$ can be written as follows:

\begin{equation}
\label{spatial}
\begin{aligned}
   L_s  &= \sum_{i=0}^{2}L_g(P^{t}_{i}, G) + \alpha\sum_{i=0}^{1}L_d(P^{t}_{i}, P^{t}_{2}),
\end{aligned}
\end{equation} where $L_g$ is the sigmoid cross entropy loss function, which computes the errors between the saliency predictions and the corresponding ground truths. $L_d$ is the distillation loss, which is the sigmoid cross entropy as well, but it computes the errors between the final saliency prediction and previous saliency predictions.

\begin{figure}[tbp]
\centering
\includegraphics[width=0.5\textwidth]{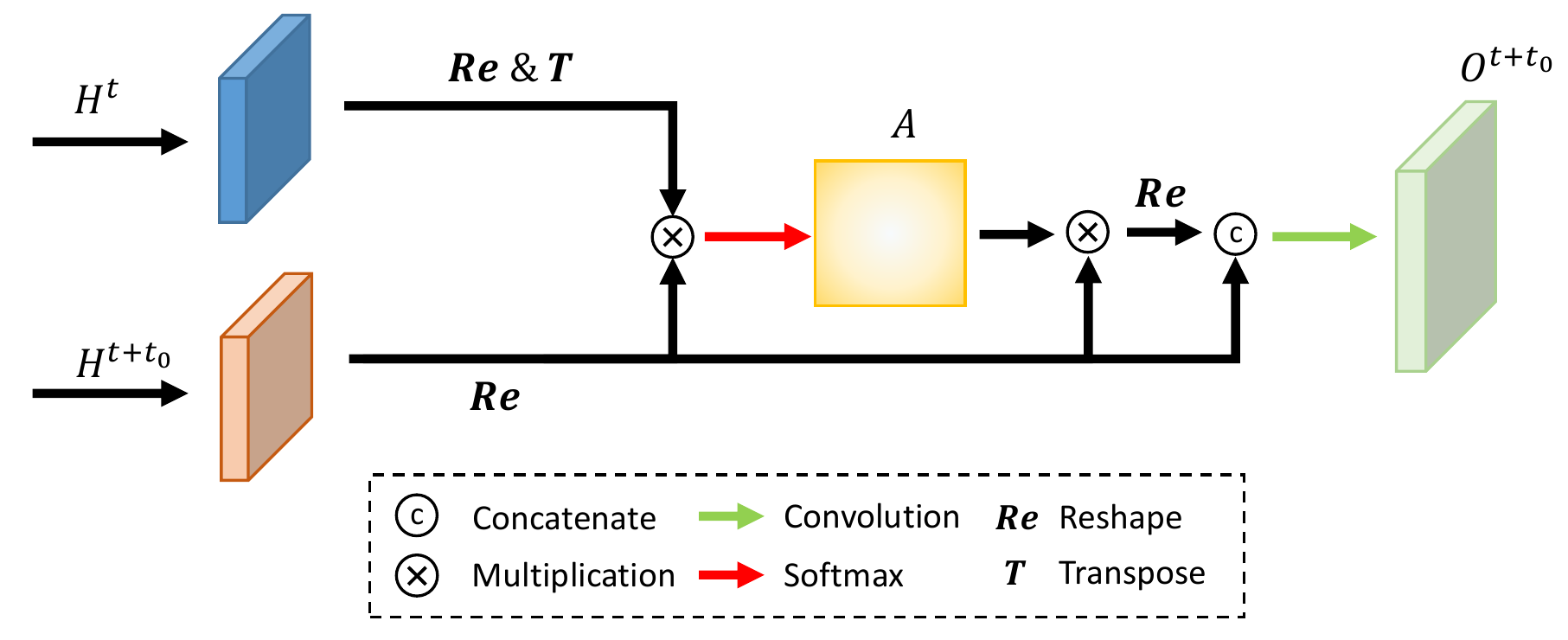}
\caption{The pipeline of the infer-frame encoding module. The high-level features from two adjacent frames are exploited to generate the weighted features by a series of operations.}
\label{attention}
\vspace{-10px}
\end{figure}

\subsection{Temporal knowledge distillation}

As for the temporal distillation, we can directly regard the generated saliency maps from adjacent frames as the supervision to propagate the sequential information. The corresponding loss function of the temporal distillation $L_t$ can be written as follow:

\begin{equation}
\label{temporal}
\begin{aligned}
   L_t  &= \sum_{i=0}^{2}L_g(P^{t+t_0}_{i}, G) + \alpha\sum_{i=0}^{2}L_d(P^{t+t_0}_{i}, P^{t}_{2}),
\end{aligned}
\end{equation}

However, as the complexity and sufficiency of the sequential information, the performance has only a slight improvement through this kind of distillation. Therefore, we consider to introduce extra sequential information to support the temporal distillation. Inspired by the attention structure \cite{vaswani2017attention,fu2018dual,yang2019anchor}, we introduce a temporary infer-frame feature encoding module, which is able to encode the weighted features generated by a mutual attention block. As shown in Fig.\ref{framework} (c). The first two feature embedding units are the same as the units in Fig.\ref{framework} (b) and their parameters are shared. As for the third feature embedding unit, its input is the weighted features from the mutual attention block, whose process is described in Fig.\ref{attention}. The generation of the weighted features can be described as below:




\begin{equation}
\label{spatial}
\begin{aligned}
   \hat{H}^{t} &= Re(H^{t}),\; \hat{H}^{t+t_0} = Re(H^{t+t_0}) \\
   A  &= \sigma(\frac{1}{\sqrt{c}} * T(\hat{H}^{t}) * \hat{H}^{t+t_0}) \\
   \tilde{H}^{t+t_0} &= \hat{H}^{t+t_0} * A \\
   O^{t+t_0}  &= Conv(Cat(Re(\tilde{H}^{t+t_0}), H^{t+t_0})),
\end{aligned}
\end{equation}

Given the high-level features $H^{t}$, $H^{t+t_0} \in R^{c \times w \times h}$ from two frames, they are firstly reshaped to $\hat{H}^{t}, \hat{H}^{t+t_0} \in R^{c \times n}$ ($n = w * h$) by the reshape operation $Re(\cdot)$. Then, $\hat{H}^{t}, \hat{H}^{t+t_0}$ are used to generate the attention map by softmax activation function $\sigma(\cdot)$ and the transpose operation $T(\cdot)$.  After that, the attention map is multiplied by $\hat{H}^{t+t_0}$ to generate the weighted feature maps $\tilde{H}^{t+t_0}$. $\tilde{H}^{t+t_0}$ is converted to original scale $R^{c \times w \times h}$. Finally, we feed the weighted feature maps and the original high-level feature maps to a convolutional layer to obtain the final fusing feature map $O^{t+t_0}$.

Since the introduction of the infer-frame encoding module, the loss function of temporal distillation can be re-written as below:

\begin{equation}
\label{temporal2}
\begin{aligned}
   L_t  &= \sum_{i=0}^{3}L_g(P^{t+t_0}_{i}, G) + \alpha\sum_{i=0}^{3}L_d(P^{t+t_0}_{i}, P^{t}_{2}),
\end{aligned}
\end{equation}

After the training stage, the sequential information is fully learned. Therefore, the temporary feature encoding module can be removed to retain the efficiency of the lightweight network. The second stage saliency prediction is our final result.

\begin{table*}[t]
\centering
\caption{Quantitative comparison results between the state-of-the-arts and the proposed approach. ``-'' denotes the dataset is used for training in their methods. The best three scores are labeled in \textcolor{red}{red}, \textcolor{blue}{blue} and \textcolor{green}{green}, orderly.}
\label{compare_all}
\begin{tabular}{|c|c|c|c|c|c|c|c|c|c|c|c|c|}
\hline
\multirow{2}{*}{Method} & \multicolumn{2}{c|}{FBMS-T} & \multicolumn{2}{c|}{DAVIS-T} & \multicolumn{2}{c|}{SegTrack-V2} & \multicolumn{2}{c|}{ViSal} & \multicolumn{2}{c|}{VOS-T} & \multicolumn{2}{c|}{DAVSOD} \\ \cline{2-13}
 & $F_\beta\uparrow$ & MAE$\downarrow$ & $F_\beta\uparrow$ & MAE$\downarrow$ & $F_\beta\uparrow$ & MAE$\downarrow$ & $F_\beta\uparrow$ & MAE$\downarrow$ & $F_\beta\uparrow$ & MAE$\downarrow$ & $F_\beta\uparrow$ & MAE$\downarrow$ \\ \hline
SIVM \cite{rahtu2010segmenting} & 0.426 & 0.236 & 0.450 & 0.212 & 0.581 & 0.251 & 0.522 & 0.197 & 0.439 & 0.217 & 0.298 & 0.288 \\ \hline
TIMP \cite{zhou2014time} & 0.456 & 0.192 & 0.488 & 0.172 & 0.573 & 0.116 & 0.479 & 0.170 & 0.401 & 0.215 & 0.395 & 0.195 \\ \hline
SPVM \cite{liu2014superpixel} & 0.330 & 0.209 & 0.390 & 0.146 & 0.618 & 0.108 & 0.700 & 0.133 & 0.351 & 0.223 & 0.358 & 0.202 \\ \hline
RWRV \cite{kim2015spatiotemporal} & 0.336 & 0.242 & 0.345 & 0.199 & 0.438 & 0.162 & 0.440 & 0.188 & 0.422 & 0.211 & 0.283 & 0.245 \\ \hline
MB \cite{zhang2015minimum} & 0.487 & 0.206 & 0.470 & 0.177 & 0.554 & 0.146 & 0.692 & 0.129 & 0.562 & 0.158 & 0.342 & 0.228 \\ \hline
SAGM \cite{wang2015saliency} & 0.564 & 0.161 & 0.515 & 0.103 & 0.634 & 0.081 & 0.688 & 0.105 & 0.482 & 0.172 & 0.370 & 0.184 \\ \hline
GFVM \cite{wang2015consistent} & 0.571 & 0.160 & 0.569 & 0.103 & 0.592 & 0.091 & 0.683 & 0.107 & 0.506 & 0.162 & 0.334 & 0.167 \\ \hline
MSTM \cite{tu2016real} & 0.500 & 0.177 & 0.429 & 0.165 & 0.526 & 0.114 & 0.673 & 0.095 & 0.567 & 0.144 & 0.344 & 0.211 \\ \hline
STBP \cite{xi2016salient} & 0.595 & 0.152 & 0.544 & 0.096 & 0.640 & 0.061 & 0.622 & 0.163 & 0.526 & 0.163 & 0.410 & 0.160 \\ \hline
SGSP \cite{liu2016saliency} & 0.630 & 0.172 & 0.655 & 0.138 & 0.673 & 0.124 & 0.677 & 0.165 & 0.426 & 0.236 & 0.426 & 0.207 \\ \hline
SFLR \cite{chen2017video} & 0.660 & 0.117 & 0.727 & 0.056 & 0.745 & 0.037 & 0.779 & 0.062 & 0.546 & 0.145 & 0.478 & 0.132 \\ \hline
SCOM \cite{Chen2018TIP} & 0.797 & 0.079 & 0.783 & 0.048 & 0.764 & 0.030 & 0.831 & 0.122 & 0.690 & 0.162 & 0.464 & 0.220 \\ \hline
SCNN \cite{8419765} & 0.762 & 0.095 & 0.714 & 0.064 & - & - & 0.831 & 0.071 & 0.609 & 0.109 & 0.532 & 0.128 \\ \hline
DLVS \cite{8047320} & 0.759 & 0.091 & 0.708 & 0.061 & - & - & 0.852 & 0.048 & 0.675 & 0.099 & 0.521 & 0.129 \\ \hline
FGRNE \cite{li2018flow} & 0.767 & 0.088 & 0.783 & 0.043 & - & - & 0.848 & 0.045 & 0.669 & 0.097 & 0.573 & \textcolor{green}{0.098} \\ \hline
MBNM \cite{li2018unsupervised} & 0.816 & \textcolor{green}{0.047} & \textcolor{green}{0.861} & 0.031 & 0.716 & 0.026 & 0.883 & \textcolor{green}{0.020} & 0.670 & 0.099 & 0.520 & 0.159 \\ \hline
PDB \cite{song2018pyramid} & 0.821 & 0.064 & 0.855 & \textcolor{blue}{0.028} & 0.800 & \textcolor{green}{0.024} & 0.888 & 0.032 & \textcolor{green}{0.742} & 0.078 & 0.572 & 0.116 \\ \hline
SSAV \cite{Fan_2019_CVPR} & \textcolor{blue}{0.865} & \textcolor{blue}{0.040} & \textcolor{green}{0.861} & \textcolor{blue}{0.028} & \textcolor{green}{0.801} & \textcolor{blue}{0.023} & \textcolor{green}{0.939} & \textcolor{green}{0.020} & \textcolor{green}{0.742} & \textcolor{green}{0.073} & \textcolor{green}{0.603} & \textcolor{green}{0.098} \\ \hline
MGA \cite{Li_2019_ICCV} & \textcolor{red}{0.910} & \textcolor{red}{0.027} & \textcolor{red}{0.902} & \textcolor{red}{0.022} & \textcolor{red}{0.868} & \textcolor{red}{0.018} & \textcolor{red}{0.947} & \textcolor{red}{0.015} & \textcolor{red}{0.762} & \textcolor{red}{0.066} & \textcolor{red}{0.646} & \textcolor{red}{0.073} \\ \hline
Ours & \textcolor{green}{0.831} & 0.055 & \textcolor{blue}{0.883} & \textcolor{red}{0.022} & \textcolor{blue}{0.847} & 0.025 & \textcolor{blue}{0.943} & \textcolor{blue}{0.018} & \textcolor{blue}{0.752} & \textcolor{blue}{0.071} & \textcolor{blue}{0.612} & \textcolor{blue}{0.084} \\ \hline
\end{tabular}
\end{table*}

\section{Experiments}

\subsection{Experimental settings}
\textbf{Datasets.} In this section, we compared the state-of-the-arts with DAVIS \cite{perazzi2016benchmark}, FBMS \cite{ochs2013segmentation}, ViSal \cite{wang2015consistent}, SegTrack-V2 \cite{li2013video}, VOS \cite{li2017benchmark} and DAVSOD \cite{Fan_2019_CVPR}, totally 6 datasets. Among them, DAVIS contains 50 video sequences. We follow the competition settings \cite{perazzi2016benchmark} and divide them into the training set (30 videos) and testing set (20 videos). The former is used for our network training, while the latter one is used for our testing for video salient object detection. FBMS contains a training set (29 videos) and a testing set (30 videos). We employ the testing set of FBMS to validate the proposed approach. ViSal and SegTrack-V2 include 17 and 13 videos, respectively. All of them are used for our experiment. For a fair comparison, we select the videos of VOS and DAVSOD by following the setting in \cite{Fan_2019_CVPR}.

\noindent \textbf{Evaluation criteria.} As we compare our method with the state-of-the-arts in video object salient detection and unsupervised video object segmentation, we use two kinds of evaluation criteria. For the video saliency, maximum F-measure and mean absolute error (MAE) are employed as the evaluation metrics. The F-measure is defined as

\begin{equation}
\label{f_measure}
\begin{aligned}
   F_{\beta} = \frac{(1 + \beta^2) \cdot Precision \cdot Recall}{\beta^2 \cdot Precision + Recall},
 \end{aligned}
\end{equation} where $\beta^2$ is set to 0.3, and $Precision$ and $Recall$ can be obtained by computing the average value of saliency maps. The MAE can be obtained by averaging the error between the saliency maps and the corresponding ground truths. The formula can be written as

\begin{equation}
\label{f_measure}
\begin{aligned}
   MAE = \frac{1}{|\mathcal{S}|} \sum_i |\mathcal{S}(p_i) - \mathcal{G}(p_i)|,
 \end{aligned}
\end{equation} where $p_i$ represents a pixel in a frame; $\mathcal{S}$ and $\mathcal{G}$ are the saliency map and the corresponding ground truth, respectively.

%
%

\noindent \textbf{Implementation details.} We implement the proposed approach by PyTorch. The training set of DUT \cite{yang2013saliency}, DAVIS \cite{perazzi2016benchmark}, DAVSOD \cite{Fan_2019_CVPR} are used for network training. We choose the ResNet-50 as the backbone, which is pre-trained by ImageNet. The dilation rates of ASPP module follow the original configuration in \cite{chen2017rethinking}.

As for the network configuration, the optimizer we adopt is SGD. Since we have two training stages, the initial learning rates are $10^{-3}$ and $10^{-4}$, which follow ``poly" adjustment policy. For each iteration, the learning rate is multiplied by $(1 - \frac{iter}{40000})^{0.9}$. Additionally, the momentum is 0.9 and 0.95, respectively. The weight decay is set to 0.0005. The batch size of the network is set to 8. For data augmentation, random cropping (crop size: 473 $\times$ 473), random rotation (10 degrees), random horizontal flipping are used as image pre-processing.

\noindent \textbf{Training and testing.} In the training phase, we firstly train the backbone network, ASPP module feature embedding modules by using the single frames. Then, we add the spatial distillation, infer-frame feature encoding module and temporal distillation into the network and finetune the pre-trained network from the first stage. After the network training, the network has already learned the robust spatiotemporal information. Therefore, the infer-frame feature encoding module is removed from the network. The saliency prediction from the second feature unit is our final result.

\subsection{Comparison with the state-of-the-arts}

We totally compared our method with 18 approaches on 6 widely used datasets. Among these approaches, SIVM \cite{rahtu2010segmenting}, TIMP \cite{zhou2014time}, SPVM \cite{liu2014superpixel}, RWRV \cite{kim2015spatiotemporal}, MB \cite{zhang2015minimum}, SAGM \cite{wang2015saliency}, GFVM \cite{wang2015consistent}, MSTM \cite{tu2016real}, STBP \cite{xi2016salient}, SGSP \cite{liu2016saliency} and SFLR \cite{chen2017video} are non-deep learning methods, while SCOM \cite{Chen2018TIP}, SCNN \cite{8419765}, DLVS \cite{8047320}, FGRNE \cite{li2018flow}, MBNM \cite{li2018unsupervised}, PDB \cite{song2018pyramid} and SSAV \cite{Fan_2019_CVPR} are deep learning based methods.

\begin{figure*}[tbp]
\centering
\includegraphics[width=1\textwidth]{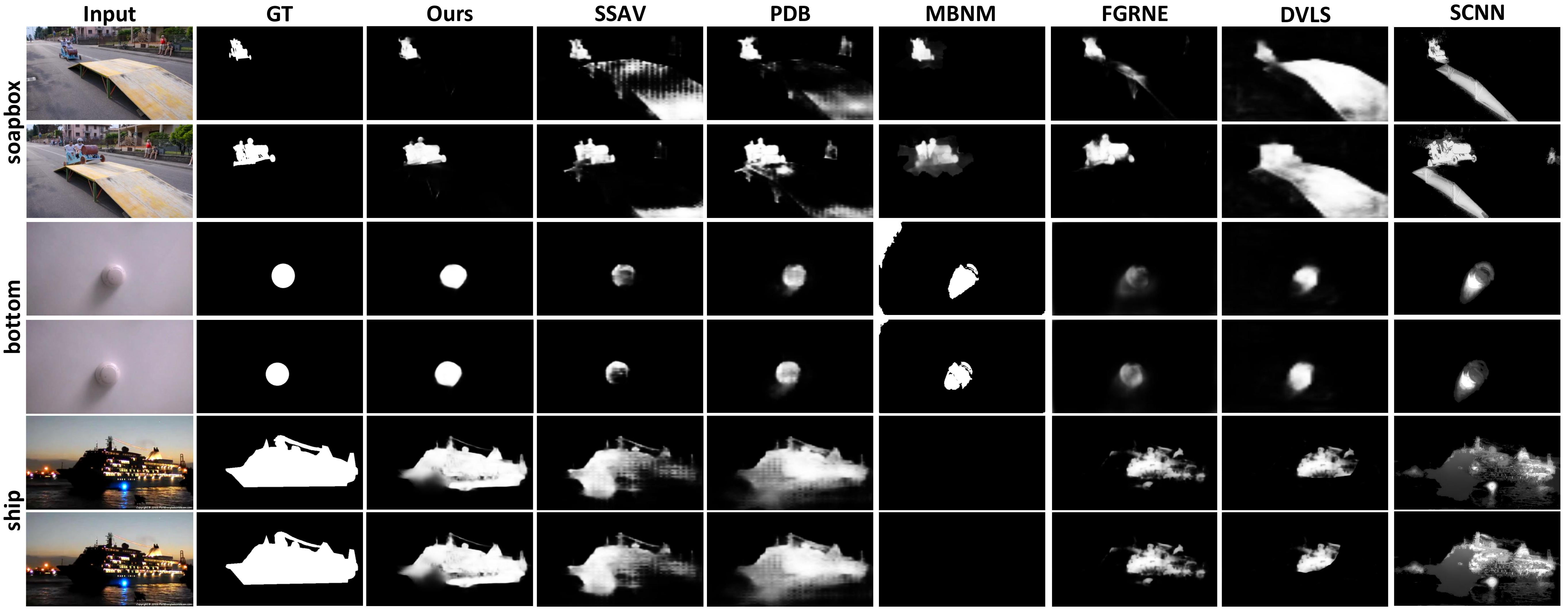}
\caption{Saliency maps generated by the deep learning-based approach and the proposed approach. Notice that the saliency maps by the proposed approach achieve competitive performance in different video scenes. }
\label{show}
\vspace{-5px}
\end{figure*}

As shown in Table.\ref{compare_all}, the proposed approach achieves competitive performance without any post-processing method. On the DAVIS and SegTrack datasets, the proposed approach has high improvement, which outperforms the SSAV approach by 2.2\% and 4.6 \% in F-measure, respectively. On the new DAVSOD and VOS dataset, our approach also boosts almost 1\% in F-measure. Besides, compared with SSAV, the proposed approach does not exploit the eye-fixation ground truths for network training. As for the ViSal, the results of the state-of-the-arts are closed, because the video length is short and the video scenes are easy. Although the proposed approach achieves the competitive performance, it cannot outperform the best one MGA, whose structure is a two-stream network. MGA proposes motion guided attention modules, which extracts the sequential features from the optical flow and then progressively fuse these features with the deep features from the video frames. The optical-based modules are very efficient, so MGA acheves the best performances in these datasets. However, the runtime of MGA is not very fast. The specific analysis is presented in the next section.

Fig.\ref{show} displays the visual comparison between the state-of-the-arts and the proposed approach. Notice that our approach can handle a variety of complex video scenes, such as high contrast regions (\textit{soapbox}), shadow disturbing (\textit{bottom}) and dark scene (\textit{ship}). For example, in the \textit{ship} sequences, the salient object is a ship in the dark night. As for this kind of scenario, the other approaches have more or less shortcomings. MBNM \cite{li2018unsupervised} lose the whole salient object. As the illumination variation, SSAV \cite{Fan_2019_CVPR} cannot detect the tail of ship. The saliency maps of PDB \cite{song2018pyramid} highlight the shadow regions of ship, which do not belong the ground truths. The proposed network obtain robust spatiotemporal features, thus detecting the entire salient object in this scene.

The reasons of the results can be concluded into two aspects. The first one is the spatial distillation. From \cite{deng2018r3net}, we know that the deeper the saliency generator is, the better saliency prediction we can obtain. We treat the deep saliency prediction as the supervision, which can revise the error of the shallow saliency prediction, thus further improve the robustness of the spatial features from the shallow layers. The second aspect is the infer-frame feature encoding and its corresponding temporal distillation, which allow the proposed network to learn the reliable sequential information. Observed by the saliency maps of consecutive frames, we find that the similar two frames can sometimes generate different results. This kind of phenomenon stems from the inadequate propagation of the sequential information. It is the proposed temporal distillation to solve the problem by the feature embedding and supervised training.

\subsection{Runtime analysis}

\begin{table}
\caption{Average runtime comparison between the proposed approach and the other video object detection approaches.}
\label{runtime}
\begin{tabular}{|c|c|c|c|c|}
\hline
Method & Ours & MGA \cite{Li_2019_ICCV} & SSAV \cite{Fan_2019_CVPR} & PDB \cite{song2018pyramid} \\ \hline
Time(s) & 0.01 & 0.04 & 0.05 & 0.05 \\ \hline
Method & MBNM \cite{li2018unsupervised} & FGRN \cite{li2018flow} & DLVS \cite{8047320} & SCNN \cite{8419765} \\ \hline
Time(s) & 2.63 & 0.09 & 0.47 & 2.53 \\ \hline
Method & SCOM \cite{Chen2018TIP} & SFLR \cite{chen2017video} & SGSP \cite{liu2016saliency} & STBP \cite{xi2016salient} \\ \hline
Time(s) & 38.8 & 119.4 & 51.7 & 49.49 \\ \hline
Method & MSTM \cite{tu2016real} & GFVM \cite{wang2015consistent} & SAGM \cite{wang2015saliency} & MB \cite{zhang2015minimum} \\ \hline
Time(s) & 0.02 & 53.7 & 45.4 & 0.02 \\ \hline
Method & RWRV \cite{kim2015spatiotemporal} & SPVM \cite{liu2014superpixel} & TIMP \cite{zhou2014time} & SIVM \cite{rahtu2010segmenting} \\ \hline
Time(s) & 18.3 & 56.1 & 69.2 & 72.4 \\ \hline
\end{tabular}
\vspace{-10px}
\end{table}

We report the runtime comparison results with 19 video object detection approaches. As shown in Table.\ref{runtime}, as the extraction of optical flow (e.g., SCOM, SCNN) or the stepwise consecutive frames feature embedding (e.g., DLVS), these methods are time-consuming. Especially for the optical flow based method, the optical flow needs to be extracted in advance. If a non-deep learning-based optical flow method are employed, the extraction of optical flow can be slower, which further affects the runtime of saliency detection. The end-to-end deep learning approaches (e.g., SSAV, PDB, FGRN) are much faster than the others. However, in order to obtain more accurate saliency results, they insert the RNN-based module (e.g., LSTM) into their models. This kind of module is effective to boost the accuracy, but directly affects the runtime of their algorithms as well. As for MGA, its network forward process is fast (0.04s per frame), because the network doesn't introduce the recurrent modules to process the multiple frames. However, MGA is not a real end-to-end method. The acquisition of optical flow needs an extra model, like FlowNet2.0 \cite{8099662}. Therefore, if the extraction process of optical flow is considered, the entire runtime of MGA should be 0.78s per frame.

In this paper, we present a lightweight network, which obtains the sequential information by the spatiotemporal distillation strategy. Therefore, the runtime of the proposed approach can achieve 0.01s per frame.

\begin{table}[t]
\caption{Ablation study on DAVIS. BS stands for the baseline scenario. SD denotes the spatial knowledge distillation; TD represents the temporal knowledge distillation; FE$_o$ demonstrates that the infer-frame feature encoding module is only used in training stage. FE$_t$ demonstrates that the infer-frame feature encoding module is used in both training and testing.}
\label{diff_compare}
\centering
\begin{tabular}{ccccc|cc}
\hline
Method  & SD     & TD     & FE$_o$ & FE$_t$ & $F_\beta$ $\uparrow$ & MAE $\downarrow$ \\ \hline
BS      &         &         &         &         & 0.842                & 0.044            \\
Ours$_k$  & $\surd$ &         &         &         & 0.861                & 0.030            \\
Ours$_t$  & $\surd$ & $\surd$ &         &         & 0.869                & 0.026            \\
Ours$_{o}$ & $\surd$ &         & $\surd$ &         & 0.872                & 0.024            \\
Ours$_{f}$ & $\surd$ & $\surd$ &  & $\surd$ & 0.881                & 0.024            \\
Ours    & $\surd$ & $\surd$ & $\surd$ &         & \textbf{0.883}       & \textbf{0.022}   \\ \hline
\end{tabular}
\vspace{-10px}
\end{table}

\begin{figure}[t]
\centering
\includegraphics[width=0.48\textwidth]{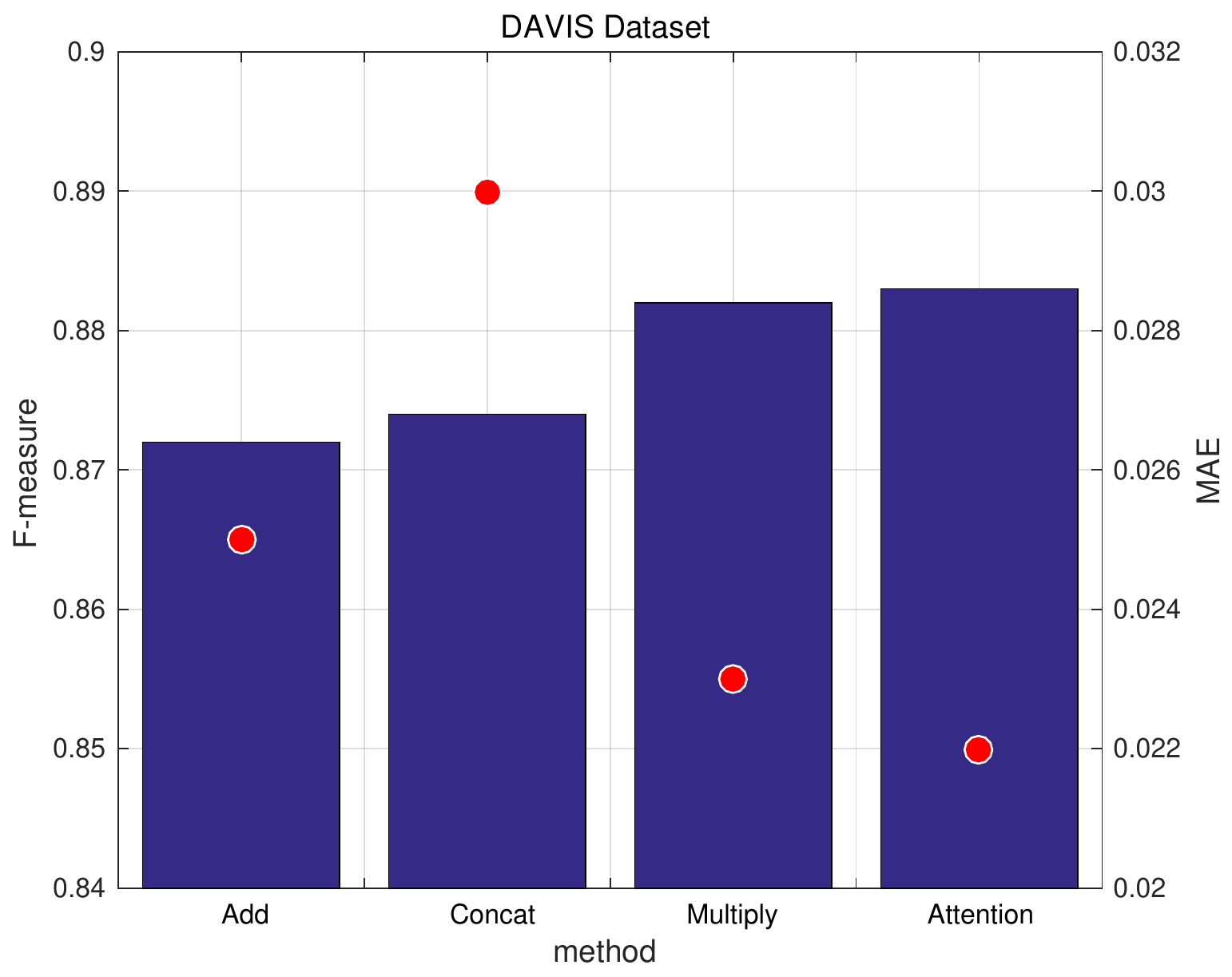}
\caption{Comparison with different infer-frame feature encoding on DAVIS dataset. The F-measures are represented by bars and the red points denote the MAEs of different weights.}
\label{infer_compare}
\vspace{-10px}
\end{figure}


\subsection{Effectiveness of different components}

In this section, we design an experiment to validate the effectiveness of different components in the proposed method. Table.\ref{diff_compare} displays the different scenarios by the proposed approach. Firstly, the baseline is tested in this experiment. Its structure contains the backbone (ResNet-50), ASPP module and saliency guidance feature embedding. Secondly, we introduce the spatial knowledge distillation into the baseline. We can see that the performance is obviously improved. The F-measure is increased by 1.9\% and the MAE is decreased by 1.4\%. After that, we directly exploit the temporal knowledge distillation to learn the sequential information, but the performance is slightly boosted. Notice that the sequential information is difficult to learn only with the supervision of adjacent frames. Therefore, we consider to introduce extra temporal features to assist the network to learn sequential information. Then, we find the weighted deep features from the mutual attention block are helpful. Table.\ref{diff_compare} shows that the performance can be significantly improved by the infer-frame feature encoding module. In the testing stage, the encoding module can be removed, which does not affect the performance and improve efficiency. The final results of F-measure and MAE are 0.883 and 0.022, respectively.

\subsection{Effectiveness of different infer-frame feature encoding}

In order to strength the effectiveness of the temporal distillation, we insert an extra feature embedding unit to encode the infer-frame features. In this section, we try to use different infer-frame encoding block with some naive fusions, such as `ADD', `Concat' and `Multiply' in Table.\ref{infer_compare}, respectively. The `ADD' operation is to fuse directly the high-level feature maps from two adjacent by element-wise addition. Similarly, the `Multiply' operation is to fuse the high-level feature maps with element-wise multiplication. The `Concat' operation is firstly to concatenate the high-level feature maps along the
depth dimension. Then, we use a convolutional layer with 1 $\times$ 1 kernel to fuse the feature maps. In the implement, the channels of the adjacent high-level feature maps are 256. In order to keep the same scale of feature maps with `ADD' and `Multiply' operation, the output channels of the convolutional layer with 1 $\times$ 1 kernel are 256 as well. Figure.\ref{infer_compare} shows that all of these operations are able to improve the performance of the proposed network. These prove that the infer-frame feature encoding can support the spatiotemporal distillation and significantly strength the propagation of sequential information. Especially, the effectiveness of `Multiply' operation and mutual attention block are better, which achieve 0.882 and 0.883 in F-measure, 0.023 and 0.022 in MAE, respectively.

\begin{figure}[tbp]
\centering
\includegraphics[width=0.485\textwidth]{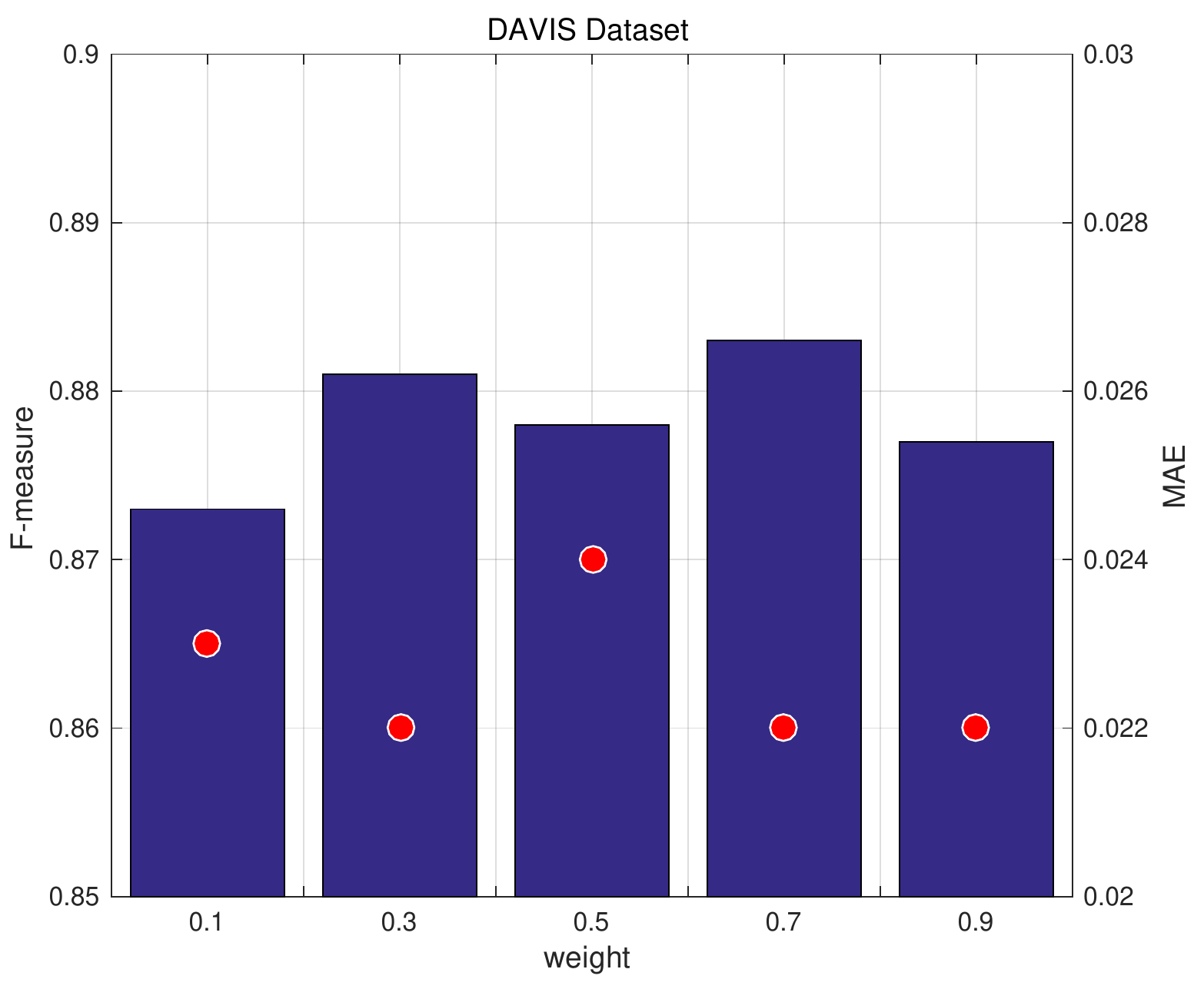}
\caption{The validation of different distillation loss weights (the value of $\alpha$) on DAVIS dataset. The F-measures are represented by bars and the red points denote the MAEs of different weights.}
\label{weights}
\vspace{-10px}
\end{figure}

\subsection{Validation of different distillation loss weights}

During the network training, we combine the original sigmoid cross entropy loss and the distillation loss to train network parameters. Therefore, we set a factor $\alpha$ to balance these two loss functions. In this section, we design an experiment to explore the most suitable value of the balance factor $\alpha$. Fig.\ref{weights} shows the results of different distillation loss weights on DAVIS dataset. The bar represents the F-measure criteria and the red points are the MAEs. As the Fig..\ref{weights} shown, we set different $\alpha$ values from 0.1 to 0.9, whose step is 0.2. The results show that the loss weights should not be too small (0.1) and too big (0.9), which directly affects the effectiveness of spatiotemporal distillation. The median (0.5) is also not the best. When the loss weight is set to 0.7, the effect of network training is the best. Therefore, the balance factor $\alpha$ is set to 0.7 in the training phase.

\section{Conclusion}
In this paper, we propose a lightweight network for video salient object detection, which learns the robust spatiotemporal features by the proposed spatiotemporal knowledge distillation strategy. The proposed distillation strategy includes two aspects. Firstly, a saliency guidance feature embedding module and the spatial distillation are combined to refine the spatial features. Secondly, we integrate an infer-frame feature encoding module and temporal distillation to effectively transfer the sequential information. As the proposed lightweight network does not exploit the complex structure to extract spatiotemporal features, it can obtain high efficiency. Additionally, the experiments also prove that our method achieves competitive performance on widely used datasets of video salient object detection.


\bibliographystyle{ACM-Reference-Format}
\bibliography{acmart}


\end{document}